\documentclass{article}

\PassOptionsToPackage{numbers,compress}{natbib}
\usepackage[preprint]{neurips_2026}

\makeatletter
\renewcommand{\@noticestring}{%
   40th Conference on Neural Information Processing Systems (NeurIPS 2026). Workshop: AI4Mat.%
}
\makeatother

\usepackage[utf8]{inputenc}
\usepackage[T1]{fontenc}
\usepackage{amsmath,amssymb,array,booktabs,graphicx,microtype,xcolor}

\usepackage{hyperref}
\hypersetup{hidelinks}
\usepackage{url}
\usepackage{placeins}
\usepackage{tabularx}

\title{Surrogate Models for Stochastic Al--Cu Phase-Field Solidification}

\author{
  Simon Savukoski \qquad Tatu Pinomaa \\
  VTT Technical Research Centre of Finland Ltd\\
  Espoo, Finland
}

\begin{document}
\maketitle

\begin{abstract}
Phase-field models resolve microstructural evolution during rapid alloy
solidification, but their computational cost limits long trajectories,
stochastic ensembles, and studies across process conditions. We compare a
deterministic U-Net with conditional flow matching (Flow), conditional
diffusion (EDM), and a variance-preserving denoising diffusion bridge
(VP-DDBM) for two-dimensional Al-Cu solidification. Given the current
phase, concentration, and prescribed temperature fields, each model
predicts the subsequent state and is applied autoregressively. At two
thermal gradients outside the training range, we compare surrogate
rollouts spanning \(48{,}000\) phase-field Euler steps with five
phase-field futures generated from a common restart at each gradient.
With two network evaluations per transition, Flow achieves the lowest
time-averaged fair continuous ranked probability score among the
generative surrogates for mean phase, mean Cu concentration, interface
length, and leading-tip position at both gradients. Flow also reproduces
the cellular morphology and retains sharp solid--liquid interfaces.
It generates a single future approximately \(107\times\) faster than
the phase-field solver on the reported hardware, showing potential
for faster process-condition screening and ensemble-based
microstructure studies.
\end{abstract}

\section{Introduction}

Phase-field (PF) models resolve diffuse solid-liquid interfaces and their
coupling to solute transport. Formulations have been developed
for both near-equilibrium alloy solidification
\citep{karma2001,echebarria2004} and rapid solidification
\citep{pinomaa2019quantitative,ji2025far}, including conditions relevant to
metal additive manufacturing \citep{ren2023solute}. However, resolving the
phase-field equations at the required spatial and temporal scales makes long
trajectories, stochastic ensembles and large parameter studies
computationally expensive.

Machine-learning surrogates can reduce the cost of PDE-governed simulation,
as demonstrated in fluid dynamics and weather forecasting
\citep{li2021fno,kochkov2021,pathak2022fourcastnet,lippe2023pderefiner}.
Convolutional, recurrent and operator-based surrogates have also been
developed for phase-field evolution
\citep{zapiain2021accelerating,bonneville2025uafno,
bonneville2025extrapolating,ji2026scalable,pinto2024leapfrog}.

In the Al--Cu simulator considered here, Gaussian noise in the phase update
influences interfacial instabilities and dendritic sidebranching
\citep{karma1999noise}. Different noise realizations can therefore produce
different microstructures from the same state under the same thermal
forcing. Probabilistic autoregressive models have been applied to stochastic
mesoscopic crystallization \citep{timmer2025crystallization}, but stochastic
surrogate prediction has received limited attention for coupled
phase-solute evolution under a moving temperature field.

We therefore formulate surrogate modeling of stochastic Al-Cu
solidification as state-conditioned forward prediction. Given the current
phase and Cu concentration fields and the prescribed temperature field, each
surrogate predicts the state 750 phase-field Euler steps later and is
applied autoregressively. We compare a deterministic U-Net with three
generative surrogates: conditional flow matching (Flow)
\citep{lipman2023flow}, conditional diffusion (EDM) \citep{karras2022edm}
and a variance-preserving denoising diffusion bridge (VP-DDBM)
\citep{zhou2024ddbm}. All four use residual U-Net backbones of approximately
matched capacity, differing mainly in the
training objective and the sampler.

Because a stochastic simulator has many valid futures, a single reference
trajectory cannot show whether a surrogate reproduces its variability. At
two thermal gradients outside the training range, we therefore compare
ensembles of surrogate rollouts with ensembles of PF futures generated from
the same restart state using independent solver noise. The initial
microstructure and the thermal forcing are thus fixed, and the PF futures
differ only through the noise. Noise can displace individual cells, so the
comparison uses physical observables and their spread across futures, not
pixel-wise agreement, together with the measured inference cost. This
comparison tests whether a generative surrogate can reproduce the
distribution of PF futures at a fraction of the solver cost.

\section{Phase-field model for rapid solidification}
At the continuum level, the dilute-alloy phase-field model follows the
formulations in Refs.~\citep{pinomaa2019quantitative,pinomaa2020phase},
which describe the co-evolution of a solid-liquid order parameter ($\phi$) and alloying element concentration field ($c$) under a prescribed temperature field $T(x,y,t)$. The order parameter equation is of the form
\begin{equation}
\begin{aligned}
 \tau(\mathbf n)\frac{\partial\phi}{\partial t}
 &=\nabla\!\cdot\!\left[W(\mathbf n)^2\nabla\phi+\mathbf f_W(\mathbf{n})\right]
   +\phi-\phi^3
   +f_D(T,\phi,c)
   +\eta_\phi, \\
\end{aligned}
\label{eq:phase}
\end{equation}
where $\mathbf n=\nabla\phi/|\nabla\phi|$ is the interface normal, $W$ and $\tau$ are anisotropic interface width and attachment time, respectively, $\mathbf f_W$ is an additional anisotropy term, $f_D(T,\phi,c)$ is the driving force term, and $\eta_\phi$ is a Gaussian stochastic forcing term following
Ref.~\citep{karma1999noise}.

The concentration follows a nonlinear diffusion equation according to
\begin{equation}
\begin{aligned}
 \frac{\partial c}{\partial t}
 &=\nabla\!\cdot\!\left[
   D_L\,c\,q(\phi)\nabla u(c,\phi)
   +
   f_{a}(\phi,c)
   \frac{\partial\phi}{\partial t}\,\mathbf n\right],
\end{aligned}
\label{eq:concentration}
\end{equation}
where $D_L$ is the solute diffusion coefficient in the liquid phase, $q(\phi)$ is an interpolation function, $u(c,\phi)$ is the chemical potential, and $f_\textrm{a}(\phi,c)$ is a prefactor to an antitrapping current used to impose controlled rapid solidification kinetics.

The imposed temperature field is prescribed as a moving linear temperature profile,
\begin{equation}
T(x,y,t)
=
T_0
+
G\left[
    x-Vt
\right],
\label{eq:thermal}
\end{equation}
where $x$ is the solidification direction coordinate, $T_0$ is an appropriate reference shift, $G$ is the imposed thermal gradient, and $V$ is the pulling speed. The PF simulator uses an explicit Euler time integration scheme with a constant time step size $\Delta t$. Details of the phase-field implementation and model parameters for the
dilute Al--Cu alloy are given in Ref.~\citep{pinomaa2025multi}.

\section{Surrogate models}
\label{sec:surrogates}

Let \(\mathbf{x}_k=[\phi_k,c_k]\) denote the phase and Cu concentration
fields at stored state \(k\), and let \(T_k\) denote the prescribed
temperature field. All surrogates model the one-step transition from
\((\mathbf{x}_k,T_k)\) to \(\mathbf{x}_{k+1}\). Temperature is prescribed
and is not predicted.

We compare a deterministic U-Net with three generative surrogates:
conditional flow matching (Flow), conditional diffusion (EDM), and a
variance-preserving denoising diffusion bridge (VP-DDBM). The deterministic
U-Net predicts \(\mathbf{x}_{k+1}\) using a residual encoder--decoder
architecture with skip connections \citep{ronneberger2015unet}. It is
trained by minimizing mean-squared error against the next stored state
and returns a single prediction. The generative surrogates instead model
a conditional distribution over possible next states.

Conditional flow matching (Flow) \citep{lipman2023flow} directly learns
a velocity field along a prescribed path from Gaussian noise to the
conditional distribution of next states. Conditional diffusion (EDM)
\citep{karras2022edm} instead learns to reconstruct the clean next state
from noise-corrupted versions at different noise levels. Its denoiser
output is then used to construct the velocity of a probability-flow ODE.
In our implementation, both samplers start from Gaussian noise and use
Euler integration. EDM returns the output of the final denoiser evaluation.

VP-DDBM models the conditional distribution using the variance-preserving
formulation of Denoising Diffusion Bridge Models \citep{zhou2024ddbm}.
During training, the known current and next states form the endpoints
of a stochastic bridge. An intermediate bridge state is sampled as
\begin{equation}
\mathbf{z}_s
=
a_s\mathbf{x}_k
+
b_s\mathbf{x}_{k+1}
+
c_s\boldsymbol{\epsilon},
\qquad
\boldsymbol{\epsilon}\sim\mathcal{N}(\mathbf{0},\mathbf{I}),
\label{eq:vp-bridge}
\end{equation}
where \(s\in[0,1]\) is the bridge time and \(a_s\), \(b_s\), and \(c_s\)
are determined by the variance-preserving schedule. The schedule satisfies
\(\mathbf{z}_1=\mathbf{x}_k\) and
\(\mathbf{z}_0=\mathbf{x}_{k+1}\), with \(c_s\) vanishing at both
endpoints. Intermediate states are therefore noisy mixtures of the current
and next states. The network is trained to recover the clean next state
from \(\mathbf{z}_s\), conditioned on \(\mathbf{x}_k\), \(T_k\), and \(s\).

At inference, \(\mathbf{x}_{k+1}\) is unknown. We use Diffusion Bridge
Implicit Models (DBIM) \citep{zheng2025dbim} to traverse the learned
bridge from the known current-state endpoint toward a sampled next state.
A first network call at \(s=1\) estimates the next state. This estimate,
the current state, and a Gaussian boot perturbation initialize the bridge
just below \(s=1\). With \(\eta=0\), the remaining DBIM updates are
deterministic given that perturbation.

All four surrogates use residual U-Net backbones with the same
temperature-conditioning design. Temperature enters through two pathways.
First, its spatial mean and standard deviation are embedded and used by
FiLM to rescale and shift feature channels, providing global thermal
conditioning \citep{perez2018film}. Second, the complete temperature field
is processed by a five-resolution convolutional encoder, and its features
are injected at the corresponding U-Net resolutions through
zero-initialized \(1\times1\) residual projections. These projections are
analogous to the zero-convolution connections used by ControlNet
\citep{zhang2023controlnet}. Unlike ControlNet, however, our architecture
does not freeze or duplicate a pretrained backbone; all backbone and
conditioning parameters are trained jointly. The generative networks also
receive the intermediate sampler state, and the global temperature
representation is combined with a sampling-time or noise-level embedding
before feature modulation. Spatial convolutions use periodic padding
along \(y\) and nonperiodic padding along \(x\), consistent with the
simulator boundary conditions.

The number of network function evaluations (NFE) counts all network
calls per transition, including the final EDM denoiser call and the
initial DBIM call. The deterministic U-Net uses one evaluation.
Appendix~\ref{sec:supp-hyperparameters} summarizes model sizes,
training hyperparameters, and sampler settings.

\section{Training and evaluation}
\label{sec:training}

The training dataset contains 70 phase-field trajectories on a uniform
\(512\times768\) grid. Simulations use a pulling speed of
\(0.2\,\mathrm{m/s}\) and 14 thermal gradients from
\(1.1\times10^{6}\) to \(2.4\times10^{6}\,\mathrm{K/m}\) in increments
of \(0.1\times10^{6}\,\mathrm{K/m}\), with five independently seeded
trajectories per gradient.

We train three networks per method on LUMI \citep{lumi_g} using
different seeds and the same optimizer, batch size, update count,
and weight-averaging scheme. Checkpoints are selected using only
the prediction error of the ensemble mean on validation states at
\(G=(0.9,\,2.5)\times10^{6}\,\mathrm{K/m}\). For U-Net, this error
is computed from its single prediction.

We evaluate autoregressive rollouts from one restart state at each
test gradient, \(G=(0.8,\,2.8)\times10^{6}\,\mathrm{K/m}\).
The phase-field solver generates five futures per restart using
independent noise. Each generative network produces five futures,
and each U-Net produces one. Each rollout spans 48,000 phase-field
Euler steps, with 750 Euler steps per surrogate transition.

We compare mean phase, domain-mean Cu concentration, interface length
\citep{tran2019quantifying}, and leading-tip position
\citep{jokisaari2018phase}, together with qualitative spatial
comparisons. For each observable, we report the fair continuous
ranked probability score (CRPS) \citep{ferro2014fair}, averaged over
64 rollout horizons. We also compare the ensemble spread of
interface length and leading-tip position. The score has the
unit of the observable and is lower for a better ensemble. For
U-Net, we report absolute error, equivalent to the CRPS of a single
deterministic prediction. At the final horizon, we also compare mean
cell spacing and its within-field standard deviation at the higher
test gradient.

For the wall-clock comparison, the phase-field solver was timed on
two CPU nodes of CSC's Roihu supercomputer \citep{csc_roihu}, each
with two 192-core AMD EPYC 9965 processors. The solver used 512 MPI
ranks, with 256 per node and one physical core per rank. Surrogate
inference was measured at batch size one on a single NVIDIA GH200
GPU with 96\,GiB of GPU memory. Solver times are medians across the
five phase-field futures and include frame output. For each network,
inference time is the median of three timed runs after one warm-up.
Reported method timings are the median over the three networks and
exclude input and output. All times are reported per surrogate
transition.
\newpage
\section{Results}
\label{sec:results}

\begin{figure}[!h]
\centering
\includegraphics[width=\linewidth]{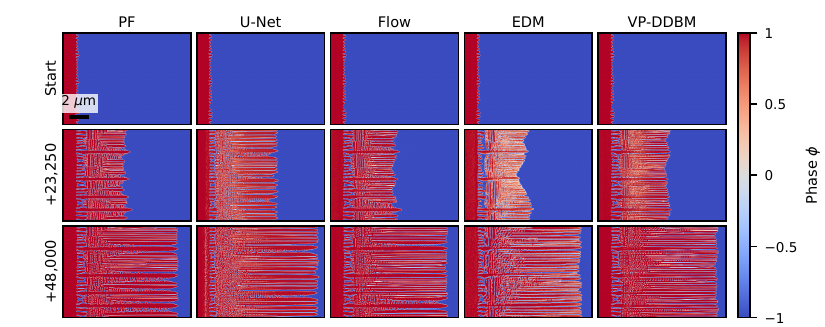}
\caption{Phase-field reference and autoregressive surrogate rollouts at
\(G=2.8\times10^{6}\,\mathrm{K/m}\). Rows show the common initial
state and the phase field \(\phi\) after \(23{,}250\) and
\(48{,}000\) phase-field Euler steps. All panels show the same crop
along the growth direction and the middle half of the transverse
domain. The generative models use two network evaluations per
transition and the U-Net uses one.}
\label{fig:phase-rollout}
\end{figure}

Figure~\ref{fig:phase-rollout} compares the phase-field reference with
one autoregressive rollout from each surrogate at the higher test
gradient. After \(23{,}250\) Euler steps, the reference front has
developed a cellular array. Flow and VP-DDBM reproduce a similar
cellular morphology. The U-Net front advances further than the
reference, while EDM produces a strongly curved front that the
reference does not show.

Figure~\ref{fig:phase-cu} shows the phase and Cu concentration fields
at both test gradients. In the rollouts shown, Flow retains sharp
solid--liquid interfaces comparable to the phase-field reference at
both displayed times. EDM produces broader interfaces, while
VP-DDBM produces broader interfaces at the intermediate horizon
but recovers sharp interfaces by the final horizon.
Figures~\ref{fig:ensemble-scores} and~\ref{fig:supp-low-gradient}
compare the rollout scores and ensemble spread at the higher and
lower test gradients, respectively. With two network evaluations
per transition, Flow achieves the lowest fair CRPS among the
generative surrogates for all four observables at both gradients,
averaged over the scored horizons. Its scores are also below the
absolute errors of the deterministic U-Net. At the higher gradient,
Flow's interface-length CRPS of \(8.3\,\mu\mathrm{m}\) is close to
the phase-field leave-one-out reference of \(7.4\,\mu\mathrm{m}\).
Its scores remain above this reference for the other observables
and at the lower gradient, so the results do not establish that
Flow reproduces the full distribution of phase-field futures.

The agreement in ensemble spread is less consistent. At the lower
gradient, Flow overestimates the spread of leading-tip position,
whereas VP-DDBM is closer to the phase-field reference. For primary
cell spacing at the higher gradient, Flow is the generative
surrogate closest to the reference in both the mean and the
within-field standard deviation at the final horizon
(Figure~\ref{fig:cell-spacing}).

For an equivalent advancement of 750 phase-field Euler steps,
the median phase-field runtime is approximately \(31.5\,\mathrm{s}\),
compared with \(0.292\)--\(0.294\,\mathrm{s}\) for Flow across the
two test gradients. On the reported hardware, this corresponds to
a wall-clock speed-up of approximately \(107\times\). The other
generative surrogates achieve similar speed-ups, while the
deterministic U-Net achieves \(211\)--\(215\times\).

At two network evaluations per transition, Flow achieves the lowest
averaged scores for all four physical observables at both test
gradients. The representative rollouts also reproduce the reference
cellular morphology and retain sharp solid--liquid interfaces
(Figure~\ref{fig:phase-cu}). Together with the reduced rollout cost,
these results make conditional flow matching a promising surrogate
for ensemble studies of microstructural evolution during rapid
Al--Cu solidification.

\clearpage
\bibliographystyle{unsrtnat}
{\small\bibliography{references}}

@article{pinomaa2020phase,
  author  = {Pinomaa, Tatu and McKeown, Joseph M. and
             Wiezorek, J{\"o}rg M. K. and Provatas, Nikolas and
             Laukkanen, Anssi and Suhonen, Tomi},
  title   = {Phase Field Modeling of Rapid Resolidification of {Al--Cu}
             Thin Films},
  journal = {Journal of Crystal Growth},
  volume  = {532},
  pages   = {125418},
  year    = {2020},
  doi     = {10.1016/j.jcrysgro.2019.125418}
}

@inproceedings{lipman2023flow,
   author = {Lipman, Yaron and Chen, Ricky T. Q. and Ben-Hamu, Heli
             and Nickel, Maximilian and Le, Matt},
   title = {Flow Matching for Generative Modeling},
   booktitle = {International Conference on Learning Representations},
   year = {2023},
   url = {https://arxiv.org/abs/2210.02747}
 }

@inproceedings{zheng2025dbim,
  author    = {Zheng, Kaiwen and He, Guande and Chen, Jianfei
               and Bao, Fan and Zhu, Jun},
  title     = {Diffusion Bridge Implicit Models},
  booktitle = {International Conference on Learning Representations},
  year      = {2025},
  url       = {https://arxiv.org/abs/2405.15885}
}

@misc{csc_roihu,
  author       = {{CSC -- IT Center for Science}},
  title        = {{Roihu} supercomputer},
  year         = {2026},
  howpublished = {\url{https://docs.csc.fi/computing/systems-roihu/}},
  note         = {Accessed 10 November 2026}
}

@inproceedings{karras2022edm,
  author = {Karras, Tero and Aittala, Miika and Aila, Timo and Laine, Samuli},
   title = {Elucidating the Design Space of Diffusion-Based Generative Models},
   booktitle = {Advances in Neural Information Processing Systems},
   volume = {35},
   pages = {26565--26577},
   year = {2022},
  url = {https://arxiv.org/abs/2206.00364}
 }

@article{ferro2014fair,
   author = {Ferro, Christopher A. T.},
   title = {Fair Scores for Ensemble Forecasts},
   journal = {Quarterly Journal of the Royal Meteorological Society},
   volume = {140},
   number = {683},
   pages = {1917--1923},
   year = {2014},
   doi = {10.1002/qj.2270}
 }

@misc{pinomaa2025multi,
  author = {Pinomaa, Tatu and Eres-Castellanos, Adriana and Thome, Pascal and
            Kaipainen, Joni and Ho, I-Ting and Hunter, Scott and Squier, Jeff and
            Fezzaa, Kamel and Biswas, Abhishek and Tin, Sammy and
            Provatas, Nikolas and Clarke, Amy J. and McKeown, Joseph and
            Voisin, Thomas},
  title  = {Multi-Scale In Situ Experiment/Simulation Integration to Predict
            Rapid Solidification Microstructures in Additively Manufactured
            {AlCu}},
  year   = {2025},
  note   = {SSRN preprint},
  doi    = {10.2139/ssrn.5402093},
  url    = {https://ssrn.com/abstract=5402093}
}

@inproceedings{li2021fno,
  author    = {Li, Zongyi and Kovachki, Nikola and Azizzadenesheli, Kamyar and
               Liu, Burigede and Bhattacharya, Kaushik and Stuart, Andrew and
               Anandkumar, Anima},
  title     = {Fourier Neural Operator for Parametric Partial Differential
               Equations},
  booktitle = {International Conference on Learning Representations},
  year      = {2021},
  eprint        = {2010.08895},
  archivePrefix = {arXiv},
  primaryClass  = {cs.LG},
  url           = {https://arxiv.org/abs/2010.08895}
}

@inproceedings{zhou2024ddbm,
  author    = {Zhou, Linqi and Lou, Aaron and Khanna, Samar and Ermon, Stefano},
  title     = {Denoising Diffusion Bridge Models},
  booktitle = {International Conference on Learning Representations},
  year      = {2024},
  eprint        = {2309.16948},
  archivePrefix = {arXiv},
  primaryClass  = {cs.CV},
  url           = {https://arxiv.org/abs/2309.16948}
}

@article{karma1999noise,
  author  = {Karma, Alain and Rappel, Wouter-Jan},
  title   = {Phase-Field Model of Dendritic Sidebranching with Thermal Noise},
  journal = {Physical Review E},
  volume  = {60},
  number  = {4},
  pages   = {3614--3625},
  year    = {1999},
  doi     = {10.1103/PhysRevE.60.3614}
}

@article{ren2023solute,
  author  = {Ren, Neng and Li, Jun and Zhang, Ruiyao and
             Panwisawas, Chinnapat and Xia, Mingxu and Dong, Hongbiao and
             Li, Jianguo},
  title   = {Solute Trapping and Non-Equilibrium Microstructure during Rapid
             Solidification of Additive Manufacturing},
  journal = {Nature Communications},
  volume  = {14},
  number  = {1},
  pages   = {7990},
  year    = {2023},
  doi     = {10.1038/s41467-023-43563-x}
}

@article{karma2001,
  author  = {Karma, Alain},
  title   = {Phase-Field Formulation for Quantitative Modeling of Alloy
             Solidification},
  journal = {Physical Review Letters},
  volume  = {87},
  number  = {11},
  pages   = {115701},
  year    = {2001},
  doi     = {10.1103/PhysRevLett.87.115701}
}

@article{echebarria2004,
  author  = {Echebarria, Blas and Folch, Roger and Karma, Alain and
             Plapp, Mathis},
  title   = {Quantitative Phase-Field Model of Alloy Solidification},
  journal = {Physical Review E},
  volume  = {70},
  number  = {6},
  pages   = {061604},
  year    = {2004},
  doi     = {10.1103/PhysRevE.70.061604}
}

@article{pinomaa2019quantitative,
  author  = {Pinomaa, Tatu and Provatas, Nikolas},
  title   = {Quantitative Phase-Field Modeling of Solute Trapping and
             Continuous Growth Kinetics in Quasi-Rapid Solidification},
  journal = {Acta Materialia},
  volume  = {168},
  pages   = {167--177},
  year    = {2019},
  doi     = {10.1016/j.actamat.2019.02.009}
}

@article{ji2025far,
  author  = {Ji, Kaihua and Zhong, Mingwang and Karma, Alain},
  title   = {Phase-Field Model of Alloy Solidification Far from Chemical
             Equilibrium at the Solid--Liquid Interface},
  journal = {Physical Review Research},
  volume  = {7},
  number  = {3},
  pages   = {033128},
  year    = {2025},
  doi     = {10.1103/vyd6-nj4h}
}

@article{zapiain2021accelerating,
  author  = {{Montes de Oca Zapiain}, David and Stewart, James A. and
             Dingreville, R{\'e}mi},
  title   = {Accelerating Phase-Field-Based Microstructure Evolution
             Predictions via Surrogate Models Trained by Machine Learning
             Methods},
  journal = {npj Computational Materials},
  volume  = {7},
  number  = {1},
  pages   = {3},
  year    = {2021},
  doi     = {10.1038/s41524-020-00471-8}
}

@article{bonneville2025uafno,
  author  = {Bonneville, Christophe and Bieberdorf, Nathan and Hegde, Arun and
             Asta, Mark and Najm, Habib N. and Capolungo, Laurent and
             Safta, Cosmin},
  title   = {Accelerating Phase Field Simulations through a Hybrid Adaptive
             {F}ourier Neural Operator with {U}-Net Backbone},
  journal = {npj Computational Materials},
  volume  = {11},
  number  = {1},
  pages   = {14},
  year    = {2025},
  doi     = {10.1038/s41524-024-01488-z}
}

@article{perez2018film,
  author  = {Perez, Ethan and Strub, Florian and de Vries, Harm and
             Dumoulin, Vincent and Courville, Aaron},
  title   = {{FiLM}: Visual Reasoning with a General Conditioning Layer},
  journal = {Proceedings of the AAAI Conference on Artificial Intelligence},
  volume  = {32},
  number  = {1},
  pages   = {3942--3951},
  year    = {2018},
  doi     = {10.1609/aaai.v32i1.11671}
}

@article{ji2026scalable,
  author  = {Ji, Kaihua and Sun, Luning and Liu, Shusen and Zhou, Fei and
             Heo, Tae Wook},
  title   = {Scalable Autoregressive Deep Surrogates for Complex
             Microstructure Dynamics},
  journal = {Machine Learning: Science and Technology},
  volume  = {7},
  number  = {4},
  pages   = {045050},
  year    = {2026},
  doi     = {10.1088/2632-2153/ae94e4}
}

@misc{bonneville2025extrapolating,
  author        = {Bonneville, Christophe and Bieberdorf, Nathan and
                   Robbe, Pieterjan and Asta, Mark and Najm, Habib N. and
                   Capolungo, Laurent and Safta, Cosmin},
  title         = {Extrapolating Phase-Field Simulations in Space and Time with
                   Purely Convolutional Architectures},
  year          = {2025},
  eprint        = {2509.20770},
  archivePrefix = {arXiv},
  primaryClass  = {cs.CE},
  note          = {NeurIPS 2025 Workshop: ML for the Physical Sciences},
  doi           = {10.48550/arXiv.2509.20770},
  url           = {https://arxiv.org/abs/2509.20770}
}

@article{tran2019quantifying,
  author  = {Tran, Anh and Liu, Dehao and Tran, Hoang A. and Wang, Yan},
  title   = {Quantifying Uncertainty in the Process--Structure Relationship
             for {Al--Cu} Solidification},
  journal = {Modelling and Simulation in Materials Science and Engineering},
  volume  = {27},
  number  = {6},
  pages   = {064005},
  year    = {2019},
  doi     = {10.1088/1361-651X/ab2690}
}

@misc{lumi_g,
  author       = {{LUMI Consortium}},
  title        = {{GPU} Nodes -- {LUMI-G}},
  year         = {2026},
  howpublished = {\url{https://docs.lumi-supercomputer.eu/hardware/lumig/}},
  note         = {Accessed 2026-08-30}
}

@article{jokisaari2018phase,
  author  = {Jokisaari, Andrea M. and Voorhees, Peter W. and
             Guyer, Jonathan E. and Warren, James A. and Heinonen, Olle G.},
  title   = {Phase Field Benchmark Problems for Dendritic Growth and Linear
             Elasticity},
  journal = {Computational Materials Science},
  volume  = {149},
  pages   = {336--347},
  year    = {2018},
  doi     = {10.1016/j.commatsci.2018.03.015}
}

@article{sun2007sharp,
  author  = {Sun, Ying and Beckermann, Christoph},
  title   = {Sharp Interface Tracking Using the Phase-Field Equation},
  journal = {Journal of Computational Physics},
  volume  = {220},
  number  = {2},
  pages   = {626--653},
  year    = {2007},
  doi     = {10.1016/j.jcp.2006.05.025}
}

@inproceedings{ronneberger2015unet,
  author    = {Ronneberger, Olaf and Fischer, Philipp and Brox, Thomas},
  title     = {{U}-Net: Convolutional Networks for Biomedical Image
               Segmentation},
  booktitle = {Medical Image Computing and Computer-Assisted Intervention --
               MICCAI 2015},
  series    = {Lecture Notes in Computer Science},
  volume    = {9351},
  pages     = {234--241},
  year      = {2015},
  publisher = {Springer},
  doi       = {10.1007/978-3-319-24574-4_28}
}

@inproceedings{zhang2023controlnet,
  author    = {Zhang, Lvmin and Rao, Anyi and Agrawala, Maneesh},
  title     = {Adding Conditional Control to Text-to-Image Diffusion Models},
  booktitle = {2023 IEEE/CVF International Conference on Computer Vision
               (ICCV)},
  pages     = {3813--3824},
  year      = {2023},
  publisher = {IEEE},
  doi       = {10.1109/ICCV51070.2023.00355}
}

@inproceedings{lippe2023pderefiner,
  author    = {Lippe, Phillip and Veeling, Bastiaan S. and Perdikaris, Paris and
               Turner, Richard E. and Brandstetter, Johannes},
  title     = {{PDE}-Refiner: Achieving Accurate Long Rollouts with Neural
               {PDE} Solvers},
  booktitle = {Advances in Neural Information Processing Systems},
  volume    = {36},
  pages     = {67398--67433},
  year      = {2023},
  url       = {https://proceedings.neurips.cc/paper_files/paper/2023/hash/d529b943af3dba734f8a7d49efcb6d09-Abstract-Conference.html}
}

@article{kochkov2021,
  author  = {Kochkov, Dmitrii and Smith, Jamie A. and Alieva, Ayya and
             Wang, Qing and Brenner, Michael P. and Hoyer, Stephan},
  title   = {Machine Learning--Accelerated Computational Fluid Dynamics},
  journal = {Proceedings of the National Academy of Sciences},
  volume  = {118},
  number  = {21},
  pages   = {e2101784118},
  year    = {2021},
  doi     = {10.1073/pnas.2101784118}
}

@misc{pathak2022fourcastnet,
  author        = {Pathak, Jaideep and Subramanian, Shashank and
                   Harrington, Peter and Raja, Sanjeev and
                   Chattopadhyay, Ashesh and Mardani, Morteza and
                   Kurth, Thorsten and Hall, David and Li, Zongyi and
                   Azizzadenesheli, Kamyar and Hassanzadeh, Pedram and
                   Kashinath, Karthik and Anandkumar, Animashree},
  title         = {{FourCastNet}: A Global Data-Driven High-Resolution Weather
                   Model Using Adaptive Fourier Neural Operators},
  year          = {2022},
  eprint        = {2202.11214},
  archivePrefix = {arXiv},
  primaryClass  = {physics.ao-ph},
  doi           = {10.48550/arXiv.2202.11214},
  url           = {https://arxiv.org/abs/2202.11214}
}

@misc{pinto2024leapfrog,
  author        = {Pinto, Damien and Greenwood, Michael and Provatas, Nikolas},
  title         = {{LeapFrog}: Getting the Jump on Multi-Scale Materials
                   Simulations Using Machine Learning},
  year          = {2024},
  eprint        = {2406.15326},
  archivePrefix = {arXiv},
  primaryClass  = {cond-mat.mtrl-sci},
  doi           = {10.48550/arXiv.2406.15326},
  url           = {https://arxiv.org/abs/2406.15326}
}

@inproceedings{timmer2025crystallization,
  author    = {Timmer, Pol and Minartz, Koen and Menkovski, Vlado},
  title     = {Efficient Probabilistic Modeling of Crystallization at
               Mesoscopic Scale},
  booktitle = {Discovery Science},
  editor    = {D{\v{z}}eroski, Sa{\v{s}}o and Levati{\'c}, Jurica and
               Pio, Gianvito and Simidjievski, Nikola},
  series    = {Lecture Notes in Computer Science},
  volume    = {16090},
  pages     = {3--16},
  publisher = {Springer},
  address   = {Cham},
  year      = {2025},
  doi       = {10.1007/978-3-032-05461-6_1}
}

\clearpage
\appendix
\numberwithin{equation}{section}

\setcounter{figure}{0}
\renewcommand{\thefigure}{S\arabic{figure}}
\setcounter{table}{0}
\renewcommand{\thetable}{S\arabic{table}}

\renewcommand{\theHfigure}{supp.\arabic{figure}}
\renewcommand{\theHtable}{supp.\arabic{table}}

\section{Supplementary materials}
\label{sec:supplementary}

\subsection{Model and training hyperparameters}
\label{sec:supp-hyperparameters}

All four surrogates were trained with AdamW, a learning rate of
\(1\times10^{-4}\), a weight decay of \(1\times10^{-2}\), and an
effective global batch size of 128. Training used exponential weight
averaging and seeds 7, 42, and 1337. Model-specific settings are
summarized in Table~\ref{tab:supp-models}.

\begin{table}[h]
  \caption{\textbf{Model-specific training and inference settings.}
  NFE denotes the total number of network evaluations per transition
  in the reported comparison, including the final EDM denoiser call
  and the initial DBIM call.}
  \label{tab:supp-models}
  \centering
  \small
  \setlength{\tabcolsep}{6pt}
  \begin{tabular}{lrc}
    \toprule
    Model & Parameters & NFE \\
    \midrule
    Deterministic U-Net
      & 229,973,858
      & 1 \\
    Flow
      & 229,859,778
      & 2 \\
    EDM
      & 229,859,778
      & 2 \\
    VP-DDBM
      & 229,859,778
      & 2 \\
    \bottomrule
  \end{tabular}
\end{table}

\FloatBarrier

Flow and EDM use Euler integration. VP-DDBM uses the DBIM sampler
with an initial Gaussian perturbation and \(\eta=0\).

Each network was trained on 16 LUMI-G nodes.

\subsection{Evaluation details}
\label{sec:evaluation-details}

Let \(R\) denote the grid points of the stored
\(512\times768\) raster, and let \(\Omega\) denote the physical
domain. The domain-mean phase and Cu concentration are defined as
\begin{equation}
\overline{\phi}(t)
=
\frac{1}{|R|}\sum_{i\in R}\phi_i(t),
\qquad
\overline{c}(t)
=
\frac{\sum_{i\in R}w_i c_i(t)}{\sum_{i\in R}w_i},
\label{eq:domain-means}
\end{equation}
where \(w_i\) is the physical-domain area represented by grid point
\(i\). The mean Cu concentration is reported in wt.\,\%.

Following the phase-field convention, the solid-liquid interface is
identified with the zero contour of the phase field
\citep{sun2007sharp}:
\begin{equation}
\Gamma_t
=
\left\{
\mathbf{x}\in\Omega:
\phi(\mathbf{x},t)=0
\right\}.
\label{eq:interface}
\end{equation}
In two dimensions, the interface length is defined as
\begin{equation}
L_{\Gamma}(t)
=
\mathcal{H}^{1}(\Gamma_t),
\label{eq:interface-length}
\end{equation}
where \(\mathcal{H}^{1}\) denotes one-dimensional Hausdorff measure.
Contour lengths are summed on the stored raster without smoothing or
joining contours across the periodic \(y\) boundary. Taking \(x\)
as the growth direction, the leading-tip position is defined as
\begin{equation}
x_{\mathrm{tip}}(t)
=
\max_{(x,y)\in\Gamma_t}x.
\label{eq:leading-tip}
\end{equation}

\begin{figure}[t]
  \centering
  \includegraphics[width=\linewidth]
    {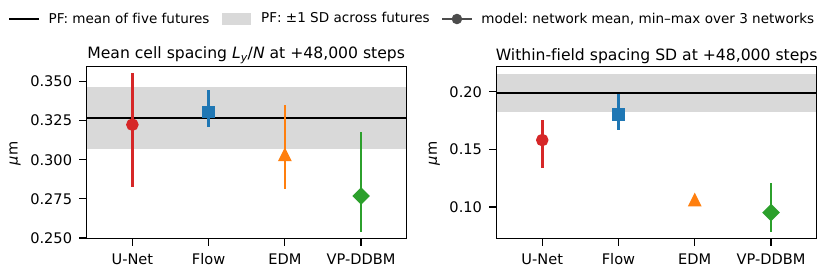}
  \caption{\textbf{Primary cell spacing at the higher test gradient.}
  The left and right panels show the mean and standard deviation
  of cell spacings within each field, respectively, after \(48{,}000\)
  phase-field Euler steps at \(G=2.8\times10^6\,\mathrm{K/m}\).
  The black line and gray band show the mean and one standard
  deviation across five phase-field futures. Markers and bars show
  the mean and range across three trained networks, with five futures
  averaged per generative network and one prediction per U-Net.
  The generative models use two network evaluations per transition
  and the U-Net uses one.}
  \label{fig:supp-cell-spacing}
  \label{fig:cell-spacing}
\end{figure}

\begin{figure}[!t]
  \centering
  \includegraphics[width=\linewidth]
    {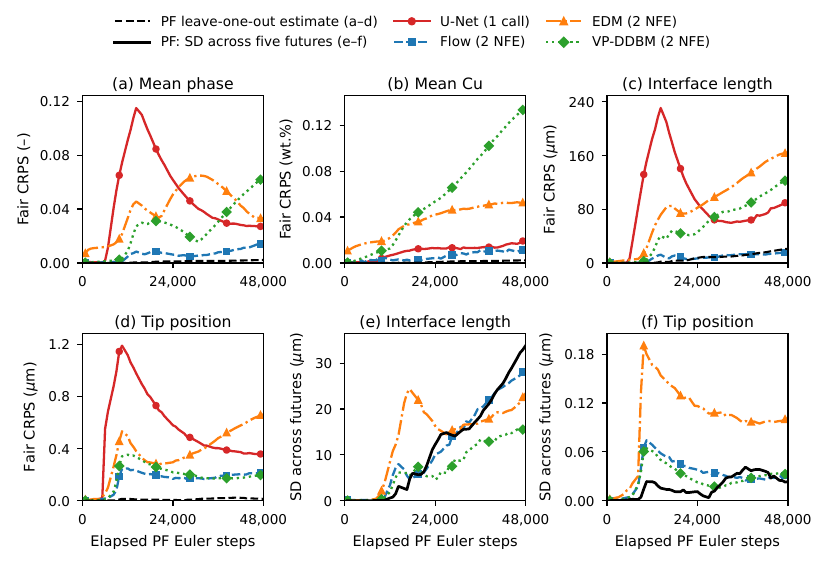}
  \caption{\textbf{Autoregressive rollout scores and ensemble spread
  at the higher test gradient.}
  Rollouts start at simulator step \(131{,}250\) at
  \(G=2.8\times10^6\,\mathrm{K/m}\) and span 64 surrogate
  transitions, corresponding to \(48{,}000\) phase-field Euler steps.
  Panels (a)--(d) show scores for mean phase, mean Cu concentration,
  interface length, and leading-tip position, respectively. Scores
  are fair CRPS for the generative models and absolute error for the
  U-Net. At each scored horizon, scores are averaged first over the
  five phase-field futures and then over the three trained networks. Panels (e) and (f)
  show the standard deviation of interface length and leading-tip
  position, respectively, across the futures of each generative
  network, averaged over the three networks. Dashed black curves
  show the phase-field leave-one-out scores; solid black curves show
  the standard deviation across the five phase-field futures.
  The generative models use two network evaluations per transition
  and the U-Net uses one.}
  \label{fig:ensemble-scores}
\end{figure}

\begin{figure}[t]
  \centering
  \includegraphics[width=\linewidth]
    {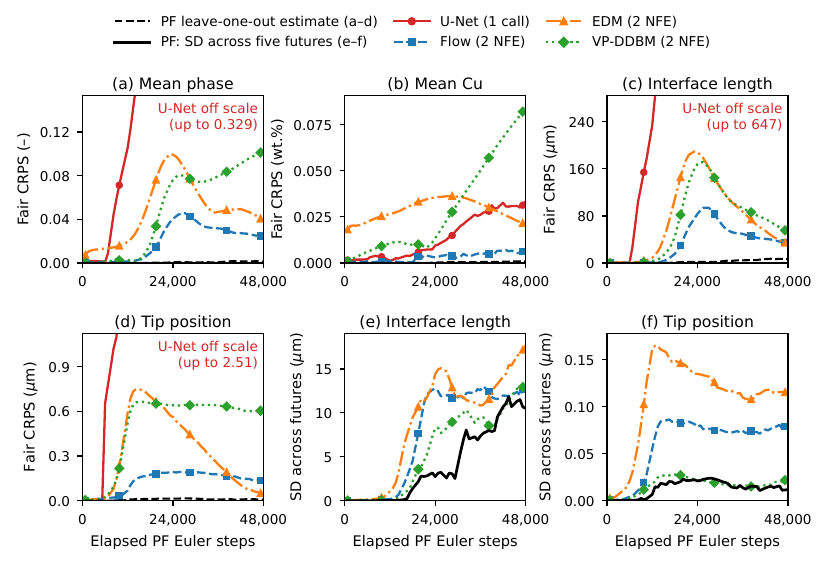}
  \caption{\textbf{Autoregressive rollout scores and ensemble spread
  at the lower test gradient.}
  Rollouts start at simulator step \(401{,}250\) at
  \(G=0.8\times10^6\,\mathrm{K/m}\) and span 64 surrogate
  transitions, corresponding to \(48{,}000\) phase-field Euler steps.
  Panels (a)--(d) show scores for mean phase, mean Cu concentration,
  interface length, and leading-tip position, respectively.
  Panels (e) and (f) show the standard deviation of interface length
  and leading-tip position across futures. Score definitions,
  averaging, and black-curve conventions are the same as in
  Figure~\ref{fig:ensemble-scores}. The generative models use two
  network evaluations per transition and the U-Net uses one.}
  \label{fig:supp-rollout-metrics-low-gradient}
  \label{fig:supp-low-gradient}
\end{figure}

\begin{figure}[t]
  \centering
  \includegraphics[width=\linewidth]
    {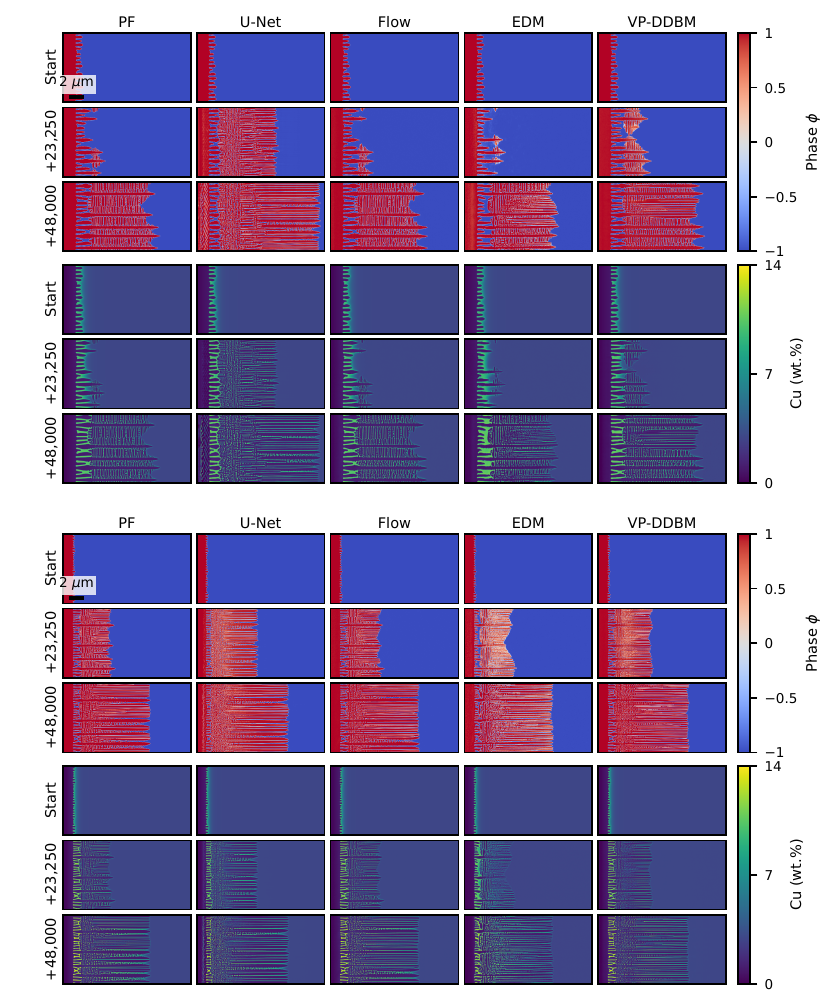}
  \caption{\textbf{Representative held-out rollouts at both test
  gradients.}
  The upper plot shows \(G=0.8\times10^6\,\mathrm{K/m}\), with
  rollouts starting at simulator step \(401{,}250\). The lower plot
  shows \(G=2.8\times10^6\,\mathrm{K/m}\), with rollouts starting
  at simulator step \(131{,}250\). Within each plot, the upper block
  shows the phase field \(\phi\) and the lower block shows Cu
  concentration in wt.\,\%. Columns show the phase-field reference,
  deterministic U-Net, Flow, EDM, and VP-DDBM. Rows within each block
  show the common initial state and the states after \(23{,}250\)
  and \(48{,}000\) phase-field Euler steps, corresponding to 31 and
  64 surrogate transitions, respectively. Each generative model
  column shows one sampled realization using two network evaluations
  per transition.}
  \label{fig:supp-all-model-rollout}
  \label{fig:phase-cu}
\end{figure}

Primary cells are detected as peaks of the solid-front profile along
the periodic \(y\) direction, using a minimum prominence of
\(0.20\,\mu\mathrm{m}\) and a minimum separation of four raster
rows. For \(N_{\mathrm{cell}}\) detected cells, the mean spacing is
\begin{equation}
\overline{\ell}
=
\frac{L_y}{N_{\mathrm{cell}}},
\qquad
L_y=18.43\,\mu\mathrm{m}.
\label{eq:cell-spacing}
\end{equation}
Figure~\ref{fig:supp-cell-spacing} compares primary cell spacing at
the higher test gradient. Figures~\ref{fig:ensemble-scores}
and~\ref{fig:supp-rollout-metrics-low-gradient} report rollout scores
and ensemble spread at the higher and lower test gradients,
respectively. Figure~\ref{fig:supp-all-model-rollout} provides a
spatial comparison of the phase and Cu concentration fields at both
gradients.

\end{document}